\documentclass{article}
\usepackage[T1]{fontenc}
\usepackage{lmodern}
\usepackage{hyperref}       
\usepackage{url}            
\usepackage{booktabs}       
\usepackage{amsfonts}       
\usepackage{nicefrac}       
\usepackage{microtype}      
\usepackage{graphicx}
\usepackage{comment}
\usepackage{url}
\usepackage{amsmath,amssymb}
\usepackage{color}
\usepackage{subfigure}
\usepackage{comment}
\usepackage{booktabs} 
\usepackage{multirow}
\usepackage{hyperref}
\usepackage{adjustbox}
\usepackage{graphicx}
\usepackage{listings}
\usepackage{diagbox}
\usepackage[normalem]{ulem}
\usepackage{wrapfig,lipsum,booktabs}
\usepackage{algorithm}
\usepackage{algorithmic}

\usepackage{hyperref}

\usepackage[accepted]{icml2021}

\begin{document}

\twocolumn[
\icmltitle{CATE: Computation-aware Neural Architecture Encoding with Transformers}



\icmlsetsymbol{equal}{*}

\begin{icmlauthorlist}
\icmlauthor{Shen Yan}{1}
\icmlauthor{Kaiqiang Song}{2,3}
\icmlauthor{Fei Liu}{2}
\icmlauthor{Mi Zhang}{1}
\end{icmlauthorlist}

\icmlaffiliation{1}{Michigan State University}
\icmlaffiliation{2}{University of Central Florida}
\icmlaffiliation{3}{Tencent AI Lab}

\icmlcorrespondingauthor{Shen Yan}{yanshen6@msu.edu}

\icmlkeywords{Machine Learning, ICML}

\vskip 0.3in
]
\printAffiliationsAndNotice{}
\begin{abstract}

Recent works \cite{white2020study,yan2020arch} demonstrate the importance of architecture encodings in Neural Architecture Search (NAS). These encodings encode either structure or computation information of the neural architectures. Compared to structure-aware encodings, computation-aware encodings map architectures with similar accuracies to the same region, which improves the downstream architecture search performance \cite{zhang2019d,white2020study}. In this work, we introduce a  \textbf{C}omputation-\textbf{A}ware \textbf{T}ransformer-based \textbf{E}ncoding method called CATE. Different from existing computation-aware encodings based on fixed transformation (e.g. path encoding), CATE employs a pairwise pre-training scheme to learn computation-aware encodings using Transformers with cross-attention. Such learned encodings contain dense and contextualized computation information of neural architectures. We compare CATE with eleven encodings under three major encoding-dependent NAS subroutines in both small and large search spaces. Our experiments show that CATE is beneficial to the downstream search, especially in the large search space. Moreover, the outside search space experiment demonstrates its superior generalization ability beyond the search space on which it was trained. Our code is available at: \href{https://github.com/MSU-MLSys-Lab/CATE}{https://github.com/MSU-MLSys-Lab/CATE}.

\end{abstract}

\section{Introduction} \label{intro}

Neural Architecture Search (NAS) has recently drawn considerable attention \cite{frank2019nassurvery}. While majority of the prior work focuses on either constructing new search spaces \cite{liu2017hierarchical,ilija2020cvpr,ru2020neural} or designing efficient architecture search and evaluation methods \cite{NAO,shi2019multiobjective,white2019bananas}, some of the most recent work \cite{white2020study,yan2020arch} sheds light on the importance of \textit{architecture encoding} on the subroutines in the NAS pipeline as well as on the overall performance of NAS.

While existing NAS methods use diverse architecture encoders such as LSTM \cite{zoph2018learning,NAO}, SRM \cite{bowen2017predictor}, MLP \cite{liu2018progressive,wang2019alphax}, GNN \cite{wen2019neural,shi2019multiobjective,yan2020arch} or adjacency matrix itself \cite{kandasamy2018neural,real2019regularized,white2020local}, these encoders encode either \textit{structures} \cite{NAO,pmlr-v97-ying19a,wang2019alphax,wen2019neural,shi2019multiobjective,yan2020arch} or \textit{computations} \cite{zhang2019d,ning2020eccv,white2019bananas} of the neural architectures. Compared to structure-aware encodings, computation-aware encodings are able to map architectures with different structures but similar accuracies to the same region. This advantage contributes to a smooth encoding space with respect to the \textit{actual} architecture performance instead of structures, which improves the efficiency of  the downstream architecture search  \cite{zhang2019d,zhang2020autobss,white2020study}.

We argue that current architecture encoders limit the power of computation-aware architecture encoding for NAS. The major limitations lie in their representation power and the effectiveness of their pre-training objectives. Specifically, \cite{zhang2019d}  uses shallow GRUs to encode computation, which is not sufficient to capture deep contextualized computation information. Moreover, their decoder is trained with the reconstruction loss via asynchronous message passing. This is very challenging in practice because directly learning the generative model based on a single architecture is not trivial. As a result, its pre-training is less effective and the downstream NAS performance is not as competitive as state-of-the-art structure-aware encoding methods. \cite{white2020study} proposes a computation-aware encoding method based on a fixed transformation called path encoding, which shows outstanding performance under the predictor-based NAS subroutine. However, path encoding scales exponentially without truncation and it inevitably causes information loss with truncation. Moreover, path encoding exhibits worse generalization performance in outside search space compared to the adjacency matrix encoding since it could not generalize to unseen paths that are not included in the training search space.

\begin{figure*}[t]
	\centering
	\includegraphics[width=0.85\textwidth]{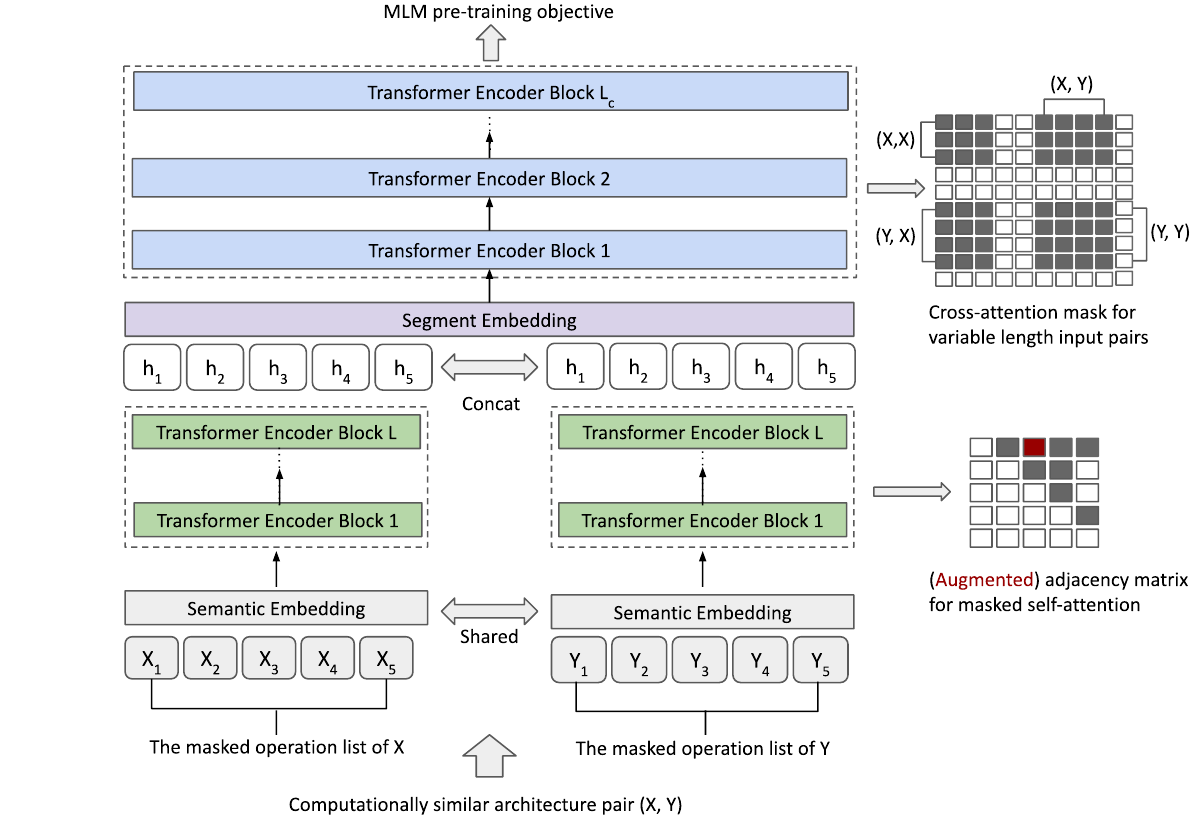}
	\caption{Overview of CATE. CATE takes computationally similar architecture pairs as the input. The model is trained to predict masked operators given the pairwise computational information. Apart from the cross-attention blocks, the pretrained Transformer encoder is used to extract architecture encodings for the downstream encoding-dependent NAS subroutines.}
	\label{fig.cate}
\end{figure*}

In this work, we propose a new computation-aware neural architecture encoding method named CATE (\textbf{C}omputation-\textbf{A}ware \textbf{T}ransformer-based \textbf{E}ncoding) that alleviates the limitations of existing computation-aware encoding methods.
As shown in Figure \ref{fig.cate}, CATE takes paired computationally similar architectures as its input.
Similar to BERT, CATE trains the Transformer-based model \cite{noam17transformer} using the masked language modeling (MLM) objective \cite{devlin2018bert}. Each input architecture pair is corrupted by replacing a fraction of their operators with a special mask token. The model is trained to predict those masked operators from the corrupted architecture pair. 

CATE differs from BERT \cite{devlin2018bert} in two aspects.
First, each prediction in LMs has its inductive bias given the contextual information from different positions. This, however, is not the case in architecture representation learning since the prediction distribution is uniform for any valid graph, making it difficult to directly learn the generative model from a single architecture. Therefore, we propose a pairwise pre-training scheme that encodes computationally similar architecture pairs  through  two  Transformers  with  shared  parameters. The two individual encodings are then concatenated, and the concatenated encoding is fed into another Transformer with a cross-attention encoder to encode the joint information of the architecture pair. Second, the fully-visible attention mask \cite{raffel2019exploring} could not be used for architecture representation learning because it does not reflect the single-directional flow (\emph{e.g.} directed, acyclic, single-in-single-out) of neural architectures \cite{Xie_2019_ICCV,You2020Graphstructure}. Therefore, instead of using a bidirectional Transformer encoder as in BERT, we directly use the adjacency matrix to compute the causal mask \cite{raffel2019exploring}. The adjacency matrix is further augmented with the Floyd algorithm \cite{floyd62shortestpath} to encode the long-range dependency of different operations. Together with the MLM objective, CATE is able to encode the computation of architectures and learn dense and deep contextualized architecture representations that contain both local and global computation information in neural architectures. This is important for architecture encodings to be generalized to outside search space beyond the training search space.

We compare CATE with eleven structure-aware and computation-aware architecture encoding methods under three major encoding-dependent subroutines as well as eight NAS algorithms on NAS-Bench-101 \cite{pmlr-v97-ying19a} (small), NAS-Bench-301 \cite{julien2020nas301} (large), and an outside search space \cite{white2020study} to evaluate the effectiveness, scalability, and generalization ability of CATE. 
Our results show that CATE is beneficial to the downstream architecture search, especially in the large search space. Specifically, we found the strongest NAS performance in all search spaces using CATE with a Bayesian optimization-based predictor subroutine together with a novel computation-aware search. Moreover, the outside search space experiment shows its superior generalization capability beyond the search space on which it was trained. Finally, our ablation studies show that the quality of CATE encodings and downstream NAS performance are non-decreasingly improved with more training architecture pairs, more cross-attention Transformer blocks and larger dimension of the feed-forward layer.  
 

\section{Related Work}
\label{sec.related}

\textbf{Neural Architecture Search (NAS).}
NAS has been started with genetic algorithms \cite{miller1989designing,kitano1990designing,ken02neat} and recently becomes popular when \cite{zoph2016neural,baker2016designing} gain significant attention. Since then, various NAS methods have been explored including sampling-based and gradient-based methods. Representative sampling-based methods include random search \cite{Li2019RandomSA}, evolutionary algorithms \cite{real2019regularized,lu2020nsganetv2}, local search \cite{ottelander2020local,white2020local}, reinforcement learning \cite{zoph2018learning,Tan2019MnasNetPN}, Bayesian optimization \cite{kandasamy2018neural,pmlr-v97-zhou19e}, Monte Carlo tree search \cite{negrinho2017deeparchitect,wang2019alphax} and Neural predictor \cite{bowen2017predictor,liu2018progressive,wen2019neural,Tang_2020_CVPR, ning2020surgery,seminas,shi2019multiobjective,yan2020arch,white2019bananas,ru2021interpretable}. Weight-sharing methods \cite{pmlr-v80-bender18a,enas} have become popular due to their computation efficiency. Based on weight-sharing, gradient-based methods are proposed to optimize the architecture selection with gradient decent \cite{NAO,liu2018darts,Xie2018SNAS,gdas,Yan2019hmnas,you2020greedynas,peng2020cream,zela2020understanding,pmlr-v119-chen20f}. For comprehensive surveys, we suggest referring to \cite{frank2019nassurvery,xie2020weight}.

\textbf{Neural Architecture Encoding.}
Majority of existing NAS work use one-hot adjacency matrix to encode the structures of neural architectures. However, adjacency matrix-based encoding grows quadratically as the search space scales up.
\cite{pmlr-v97-ying19a} proposes categorical adjacency matrix-based encoding to ensure fixed length encodings. They also propose continuous adjacency matrix-based encoding that is similar to DARTS \cite{liu2018darts}, where the architecture is created by taking fixed number of edges with the highest continuous values. 
However, this approach is not easily applicable to some NAS algorithms such as regularized evolution \cite{pmlr-v70-real17a} without major changes. 
Tabular encoding in the form of ConfigSpace \cite{lindauer2019boah} is often used for hyperparameter optimization \cite{hyperband,bohb} and recently adopted by NAS-Bench-301 \cite{julien2020nas301} to represent architectures by introducing categorical hyperparameters for each operation along each potential edge.  
Recent NAS methods \cite{luo2018neural,wang2019alphax,wen2019neural,shi2019multiobjective} use adjacency matrix as the input to LSTM/MLP/GNN to encode the structures of neural architectures in the latent space. \cite{yan2020arch} validates that pre-training architecture representations without using accuracies can better preserve the local structural relationship of neural architectures in the latent space. \cite{wei2020selfsupervised} proposes to learn architecture representations using contrastive learning to find low-dimensional embeddings. \cite{choi2021pretraining} studies various locality-based self-supervised objectives on the effect of architecture representations. One disadvantage of these methods is that they rely on a prior where the edit distance closeness between different architectures is a good indicator of the relative performance; however, structure-aware encodings may not be computationally unique unless some certain graph hashing is applied \cite{pmlr-v97-ying19a,ning2020eccv}.
\cite{white2019bananas,Wei2020NPENASNP} use path encoding and its categorical and continuous variants, which encode computation of architectures so that isomorphic cells are mapped to the same encoding. \cite{zhang2019d} uses GRU-based asynchronous message passing to encode computation of architectures and the model is trained with the VAE loss. \cite{lukasik2020smooth} proposes a two-sided variational encoder-decoder GNN to learn smooth embeddings in various NAS search spaces.
CATE is inspired by the advantage of computation encoding and addresses the drawbacks of  \cite{zhang2019d,white2019bananas}. Another line of work is based on the intrinsic properties of the architectures. \cite{hesslow2021contrastive} generates architecture representations by using contrastive learning over data Jacobian matrix values computed based on different initializations, and the generated embeddings are independent of the parameterization of the search space.

\textbf{Context Dependency.} 
Our work is close to self-supervised learning in language models (LMs) \cite{dong2019unified}. In particular, ELMo \cite{peters2018elmo} uses two shallow unidirectional LSTMs \cite{hochreiter1997long} to encode bidirectional text information, which is not sufficient for modeling deep interactions between the two directions. GPT-2 \cite{radford2019language} proposes an autoregressive language modeling method with Transformer \cite{noam17transformer} to cover the left-to-right dependency and is further generalized by XLNet \cite{xlnet19} which encodes bidirectional context. (Ro)BERT/BART/T5 \cite{devlin2018bert,liu2019robert,Lewis_2020,raffel2019exploring} use bidirectional Transformer encoder to encode both left and right context. 
In architecture representation learning, however, the attention mask in the encoder cannot be used to attend to all the operators because it does not reflect the single-directional flow of the computational graphs \cite{Xie_2019_ICCV,You2020Graphstructure}.

\section{CATE}
\label{sec.cate}

\newcommand\numberthis{\addtocounter{equation}{1}\tag{\theequation}}


\subsection{Search Space}
We restrict our search space to the cell-based architectures. Following the configuration in \cite{pmlr-v97-ying19a}, each cell is a labeled directed acyclic graph (DAG) $\mathcal{G}=(\mathcal{V}, \mathcal{E})$, with $\mathcal{V}$ as a set of $N$ nodes and $\mathcal{E}$ as a set of edges that connect the nodes. Each node $v_i \in \mathcal{V}$, $i \in [1, N]$ is associated with an operation selected from a predefined set of $V$ operations, and the edges between different nodes are represented as an upper triangular binary adjacency matrix $\mathbf{A}\in\{0, 1\}^{N\times N}$. 

\subsection{Computation-aware Neural Architecture Encoder} \label{enc}
Our proposed computation-aware neural architecture encoder is built upon the Transformer encoder architecture which consists of a semantic embedding layer and $L$ Transformer blocks stacked on top. Given $\mathcal{G}$, each operation $v_i$ is first fed into a semantic embedding layer of size $d_e$:
\begin{align}
    \mathbf{Emb}_i = \mathbf{Embedding}(v_i)
    \label{eq:emb}
\end{align}

The embedded vectors are then contextualized at different levels of abstract.
We denote the hidden state after $l$-th layer as $\mathbf{H}^l=[\mathbf{H}_1^l, ... ,\mathbf{H}_N^l]$ of size $d_h$, where $\mathbf{H}^l = T(\mathbf{H}^{l-1})$ and T is a transformer block containing $n_{head}$ heads.
The $l$-th Transformer block is calculated as:
\begin{equation}
    \small
    \mathbf{Q}_k = \mathbf{H}^{l-1}\mathbf{W}_{qk}^{l},
    \mathbf{K}_k = \mathbf{H}^{l-1}\mathbf{W}_{kk}^{l},
    \mathbf{V}_k = \mathbf{H}^{l-1}\mathbf{W}_{vk}^{l}
\end{equation}
\begin{align}
    \mathbf{\hat{H}}^l_k &= \textit{softmax}(\dfrac{\mathbf{Q}_k\mathbf{K}_k^T}{\sqrt{d_{h}}} + \mathbf{M})\mathbf{V}_k\\
    \mathbf{\hat{H}}^{l} &= \textit{concatenate}(\mathbf{\hat{H}}^l_1, \mathbf{\hat{H}}^l_2, \dots, \mathbf{\hat{H}}^l_{n_{head}})\\
    \mathbf{H}^l &= \mathbf{ReLU}(\mathbf{\hat{H}}^l\mathbf{W}_1 + \mathbf{b}_1)\mathbf{W}_2 + \mathbf{b}_2
\end{align}

where the initial hidden state $\mathbf{H}^0_i$ is $\mathbf{Emb}_i$, thus $d_e=d_h$. $\mathbf{Q}_k$, $\mathbf{K}_k$, $\mathbf{V}_k$ stand for ``Query", ``Key" and ``Value" in the attention operation of the $k$-th head respectively. \textbf{M} is the attention mask in the Transformer, where \textbf{M}$_{i,j} \in \{0, -\infty \}$ indicates whether operation $j$ is a dependent operation of operation $i$. 
$\mathbf{W}_1 \in \mathbb{R}^{d_c \times d_{ff}}$ and $\mathbf{W}_2 \in \mathbb{R}^{d_{ff} \times d_c}$ denote the weights in the feed-forward layer.

\noindent\textbf{Direct/Indirect Dependency Mask.}
A pair of nodes (operations) within an architecture are dependent if there is either a directed edge that directly connects them (\textit{local dependency}) or a path made of a series of such edges that indirectly connects them (\textit{long-range dependency}). 
We create dependency masks for such pairs of nodes for both direct and indirect cases and use these dependency masks as the attention masks in the Transformer. 
Specifically, the direct dependency mask $\mathbf{M}^{Direct}$ and the indirect dependency mask $\mathbf{M}^{Indirect}$ can be created as follows:
\begin{align*}
        \mathbf{M}^{Direct}_{i,j} &= \Big\{
    \begin{array}{lr}
        0,& \quad \textit{if} \quad A_{i,j} = 1\\
        -\infty,& \quad \textit{if} \quad A_{i,j} = 0\\
    \end{array}
     \\
    \mathbf{M}^{Indirect}_{i,j} &= \Big\{
    \begin{array}{lr}
        0,& \quad \textit{if} \quad \tilde{A}_{i,j} = 1\\
        -\infty,& \quad \textit{if} \quad \tilde{A}_{i,j} = 0\\
    \end{array}
\end{align*}

where $\mathbf{A}$ is the adjacency matrix and  $\mathbf{\tilde{A}} = \textit{Floyed}(\mathbf{A})$ is derived using Floyd algorithm in Algorithm \ref{alg:floyd}. 

\begin{algorithm}[t]
\caption{Floyd Algorithm} \label{alg:floyd}
\begin{algorithmic}[1]
    \STATE {\bfseries Input:} the node set $\mathcal{V}$, the adjacent matrix $\mathbf{A}$
    \STATE $\mathbf{\tilde{A}}  \gets \mathbf{A}$
    \FOR{$k \in \mathcal{V}$}
        \FOR{$i \in \mathcal{V}$}
            \FOR{$j \in \mathcal{V}$}
                \STATE $\mathbf{\tilde{A}}_{i,j} \quad |= \mathbf{\tilde{A}}_{i,k}  \quad \& \quad \mathbf{\tilde{A}}_{k,j}$
            \ENDFOR
        \ENDFOR
    \ENDFOR
    \STATE{\bfseries Output:} $\mathbf{\tilde{A}}$
\end{algorithmic}
\end{algorithm}

\noindent\textbf{Uni/Bidirectional Encoding.} Finally, the final hidden vector $\mathbf{H}_N^l$ is used as the unidirectional encoding for the architecture. %
We also considered encoding the architecture in a bidirectional manner, where both the output node hidden vector from the original DAG and the input node hidden vector from the reversed one are extracted and then concatenated together. However, our experiments show that bidirectional encoding performs worse than unidirectional encoding. We include this result in Appendix \ref{uni-bi-enc}.

\subsection{Pre-training CATE} \label{pretraining}

\noindent\textbf{Architecture Pair Sampling.}
We split the dataset into 95\% training and 5\% held-out test sets for our pairwise pre-training. To ensure that it does not scale with quadratic time complexity, we first sort the architectures based on their computational attributes $\mathbf{P}$ (\emph{e.g.} number of parameters, FLOPs). We then employ a sliding window for each architecture $x^i$ and its neighborhood $r(x^i) = \{y: |\mathbf{P}(x^i) - \mathbf{P}(y)| < \delta\}$, where $\delta$ is a hyperparameter for the pairwise computation constraint. Finally, we randomly select $K$ distinct architectures $Y=\{y^1, \dots, y^K\}, x^i\notin Y, Y\subset r(x^i)$ within the neighborhood to compose $K$ architecture pairs $\{ (x^i, y^1), \dots, (x^i, y^K)\}$ for architecture $x^i$.

\textbf{Pairwise Pre-training with Cross-Attention.} Once the computationally similar architecture pair is composed, 
we randomly select 20\% operations from each architecture within the pair for masking, where 80\% of them are replaced with a $[MASK]$ token and the remaining 20\% are replaced with a random token chosen from the predefined operation set. We apply padding to architectures that have nodes less than the maximum number of nodes $N$ in one batch to handle variable length inputs. The joint representation $\mathbf{H}^{L}_{XY}$ is derived by concatenating $\mathbf{H}^L_X$ and $\mathbf{H}^L_Y$ followed by the summation of the corresponding segment embedding. Segment embedding acts as an identifier of different architectures during pre-training. We set it to be trainable and randomly initialized. The joint representation $\mathbf{H}^{L}_{XY}$ is then contextualized with another $L_c$-layer Transformer with the cross-attention mask $\mathbf{M}_c$ such that segments from 
the two architectures can attend to each other given the pairwise information.  For example, given two architectures $X$ with three nodes and $Y$ with four nodes in Figure \ref{fig.cate}, $X$ has access to the non-padded nodes of $Y$ and itself, and same for $Y$.  The cross-attention dimension of the encoder is denoted as $d_c$. 
The joint representation of the last layer is used for prediction. The model is trained by minimizing the cross-entropy loss computed using the predicted operations and the original operations.

\subsection{Encoding-dependent NAS Subroutines} \label{application}

\cite{white2020study} identifies three major encoding-dependent subroutines included in existing NAS algorithms: \emph{sample random architecture}, \emph{perturb architecture}, and \emph{train predictor model}. The \emph{sample random architecture} subroutine includes random search \cite{Li2019RandomSA}. The \emph{perturb architecture} subroutine includes regularized evolution (REA) \cite{real2019regularized} and local search (LS) \cite{white2020local}. The \emph{train predictor model} subroutine includes neural predictor \cite{wen2019neural,shi2019multiobjective,white2019bananas}, Bayesian optimization with Gaussian process (GP) \cite{rasmussen2006gp}, and Bayesian optimization with neural networks (DNGO) \cite{snoek2015scalable} which is much faster to fit compared to GP and scales linearly with large datasets rather than cubically.

Inspired by \cite{ottelander2020local,white2020local}, we found that LS (\emph{perturb architecture}) can be combined with DNGO (\emph{train predictor model}). We thus propose a DNGO-based computation-aware search using CATE called CATE-DNGO-LS. 
%
Specifically, we maintain a pool of sampled architectures and take iterations to add new ones.
In each iteration, we pass all architecture encodings to the predictor trained 30 epochs with samples in the current pool.
We select new architectures with top-5 predicted accuracy and add them to the pool.
Assume there are M new architectures which become the new top-5 in the updated pool. We then select the nearest neighbors of the other (5-M) top-5 architectures in L2 distance in latent space and add them to the pool.
Hence, there will be 5 to 10 new architectures added to the pool in each iteration.
The search stops when the number of samples reaches a pre-defined budget.


\section{Experiments}
\label{sec.experiments}

\begin{figure*}[t]
	\centering
	\includegraphics[width=1.0\textwidth]{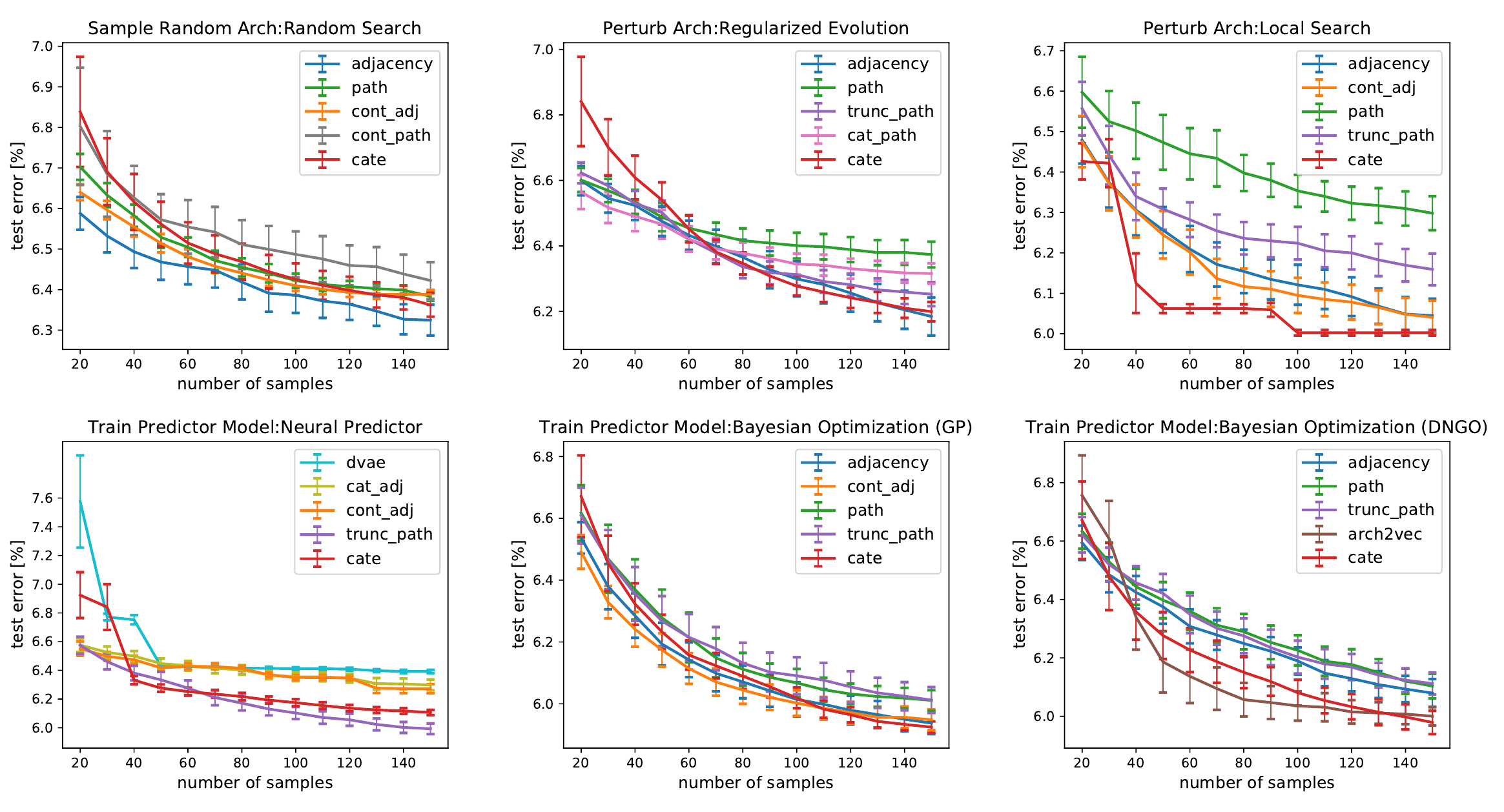}
	\caption{Comparison between CATE and other architecture encoding schemes under different subroutines on NAS-Bench-101: \emph{sample random architecture} (top left), \emph{perturb architecture} (top middle, top right), and \emph{train predictor model}
(bottom left, bottom middle, bottom right). It reports the test error of $200$ independent runs given $150$ queried architectures.}
	\label{fig.nas_encodings_comprison}
\end{figure*}

\begin{figure*}[t]
	\centering
	\includegraphics[width=1.0\textwidth]{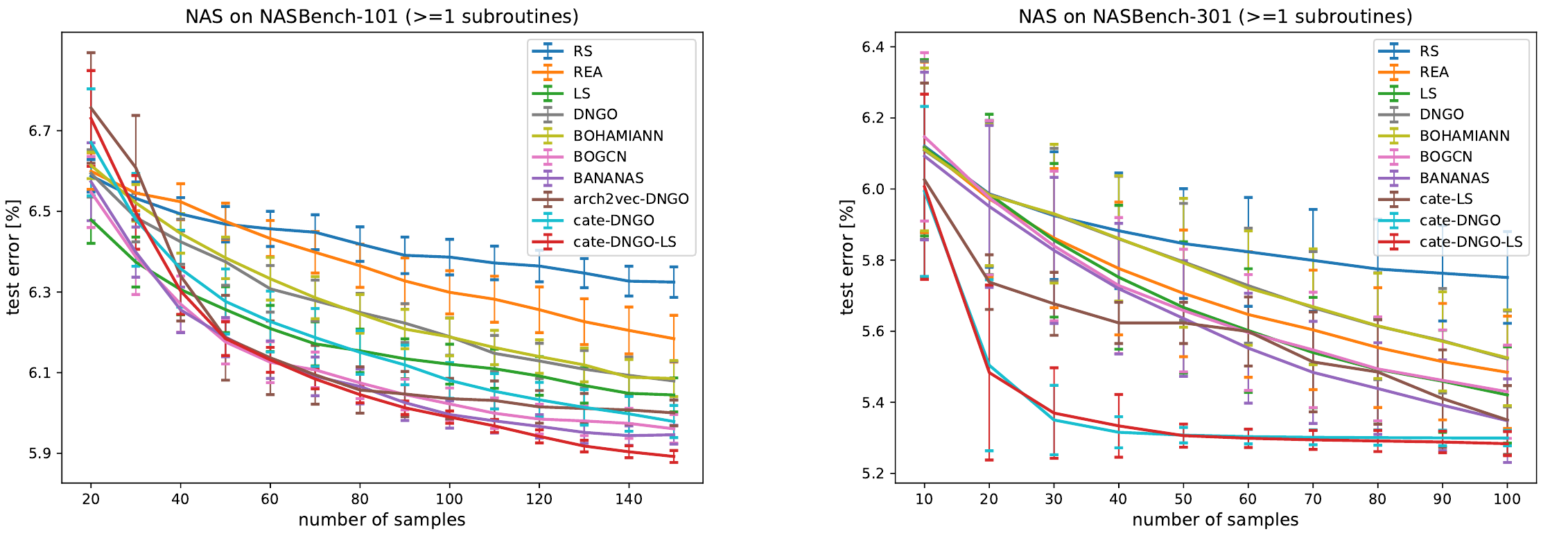}
	\caption{Comparison between CATE and SOTA NAS methods on NAS-Bench-101 (left) and NAS-Bench-301 (right). It reports the test error of $200$ independent runs. The error bars denote the variance of the test error. The number of queried architectures is set to $150$ for NAS-Bench-101 and $100$ for NAS-Bench-301.}
	\label{fig.nas_sota_comprison}
\end{figure*}

We describe two NAS benchmarks used in our experiments.

\textbf{NAS-Bench-101.}
The NAS-Bench-101 search space \cite{pmlr-v97-ying19a} consists of  $423,624$ architectures. Each architecture has its pre-computed validation and test accuracies on CIFAR-10. The cell includes up to $7$ nodes and at most $9$ edges with the first node as input and the last node as output. The intermediate nodes can be either 1$\times$1 convolution, 3$\times$3 convolution, or 3$\times$3 max pooling. We use the number of network parameters as the computational attribute $\mathbf{P}$ for architecture pair sampling. We set $\delta$ to $2,000,000$ and $K$ to 2. The ablation studies on $\delta$ and $K$ are summarized in Section \ref{sec:ablation}. We split the dataset into 95\% training and 5\% held-out test sets for pre-training.

\textbf{NAS-Bench-301.}
NAS-Bench-301 \cite{julien2020nas301} is a new surrogate benchmark on the DARTS \cite{liu2018darts} search space that is much larger than NAS-Bench-101. It 
was created by fully training $60,000$ architectures that is \emph{stratified by the NAS methods}\footnote{We suggest referring to C.2 in ~\citep{julien2020nas301} for a detailed description on the data collection.} with a good coverage and then fitting a surrogate model that can estimate the accuracy (with noise) at epoch $100$ and the training time for any of the remaining $10^{18}$ architectures.
To convert the DARTS search space into one with the same input format as NAS-Bench-101, we add a summation node to make nodes represent operations and edges represent data flow. Following \cite{liu2018progressive}, we use the same cell for both normal and reduction cell, allowing roughly $10^9$ DAGs without considering graph isomorphism. More details about the DARTS/NAS-Bench-301 and a cell transformation example are included in Appendix \ref{sec.301-cell}. 
We randomly sample $1,000,000$ architectures in this search space, and use the same data split used in NAS-Bench-101 for pre-training. We use network FLOPs as the computational attribute $\mathbf{P}$ for architecture pair sampling. We set $\delta$ to $5,000,000$ and $K$ to 1. %
Since some NAS methods we compare against use the same GIN \cite{xu2018how} surrogate model used in NAS-Bench-301, to ensure fair comparison, we thus followed \cite{julien2020nas301} to use XGB-v1.0 and LGB-runtime-v1.0 which utilizes gradient boosted trees \cite{Chen:2016:XST:2939672.2939785,NIPS2017_6449f44a} as the regression model. 

\textbf{Model and Training.}
We use a $L = 12$ layer Transformer encoder and a $L_c = 24$ layer cross-attention Transformer encoder, each has 8 attention heads. The hidden state size is $d_h =d_c = 64$ for all the encoders. The hidden dimension is $d_{ff} = 64$ for all the feed-forward layers.
We employ AdamW \cite{loshchilov2017decoupled} as our optimizer.
The initial learning rate is 1e-3.
The momentum parameters are set to 0.9 and 0.999.
The weight decay is 0.01 for regular layer and 0 for dropout and layer normalization.
We trained our model with batch size of 1024 on NVIDIA Quadro RTX 8000 GPUs. It takes around 4GB GPU memory for NAS-Bench-101 and 9GB GPU memory for NAS-Bench-301. The validation loss converges well after 10 epochs of pretraining, which takes 1.2 hours on NAS-Bench-101 and 7.5 hours on NAS-Bench-301. 

\subsection{Comparison with Different Encoding Schemes} \label{enc:comp}
%
In our first experiment, we compare CATE with eleven architecture encoding schemes under three major encoding-dependent subroutines described in Section \ref{application} on NAS-Bench-101.
These encoding schemes include (1-3) one-hot/categorical/continuous adjacency matrix encoding  \cite{pmlr-v97-ying19a},  (4-6) one-hot/categorical/continuous path encoding and (7-9) their corresponding truncated counterparts \cite{white2019bananas}, (10) D-VAE \cite{zhang2019d}, and (11) arch2vec \cite{yan2020arch}. For continuous encodings, we use L2 distance as the distance metric. To examine the effectiveness of the encoding schemes themselves, we compare different encoding schemes under the same search subroutine. 

Figure \ref{fig.nas_encodings_comprison} illustrates our results. For each subroutine, we show the top-five best-performing encoding schemes. Overall, despite there is no overall best encoding, we found that CATE is among the top five across all the subroutines.

Specifically, for \emph{sample random architecture} subroutine, random search using adjacency matrix encoding performs the best. 
The random search using continuous encodings performs slightly worse than the adjacency encodings possibly due to the discretization loss from vector space into a fixed number of bins of same size before the random sampling. 

For \emph{perturb architecture} subroutine, CATE is on par with or outperforms adjacency encoding and path encoding because it is pre-trained to preserve strong computation locality information. This advantage allows the evolution or local search to find architectures with similar performance in local neighborhood more easily. Interestingly, we observe very small deviation using local search with CATE. This indicates that it always converges to some certain local minimums across different initial seeds. Since NAS-Bench-101 already exhibits locality in edit distance, encoding computation makes architectures even closer in terms of accuracy and thus benefits the local search. 

For \emph{train predictor model} subroutine, we have four observations: 
1) Adjacency matrix encodings perform less effective with neural predictor and DNGO. It is possibly that edit distance cannot fully reflect the closeness of architectures w.r.t their actual performance. 
2) Path encoding performs well with neural predictor but worse than other encodings with Bayesian optimization.
3) D-VAE and arch2vec, two encodings learned via variational autoencoding, perform well only with some certain NAS methods. It could be attributed to their challenging training objective which easily leads to overfitting. 
4) CATE is competitive with neural predictor and outperforms all the other encodings with Bayesian optimization. This is because neighboring computation-aware encodings correspond with similar accuracies. Moreover, the training objective in CATE is more efficient compared to the standard VAE loss \cite{kingma2013autoencoding} used by D-VAE and arch2vec.

\subsection{Comparison with Different NAS Methods}

In our second experiment, we compare the neural architecture search performance based on CATE encodings with state-of-the-art NAS algorithms on NAS-Bench-101 and NAS-Bench-301. Existing NAS algorithms contain one or more encoding-dependent subroutines. 
We consider six NAS algorithms that contain one encoding-dependent subroutine: random search (RS) \cite{Li2019RandomSA} (\emph{sample random arch.}), regularized evolution (REA) \cite{real2019regularized} (\emph{perturb arch.}), local search (LS) \cite{white2020local} (\emph{perturb arch.}), DNGO \cite{snoek2015scalable} (\emph{train predictor}), BOHAMIANN \cite{NIPS2016_springenberg} (\emph{train predictor}), arch2vec-DNGO \cite{yan2020arch} (\emph{train predictor}), and two NAS algorithms that contain more than one encoding-dependent subroutine: BOGCN \cite{shi2019multiobjective} (\emph{perturb arch.}, \emph{train predictor}) and BANANAS \cite{white2019bananas} (\emph{sample random arch.}, \emph{perturb arch.}, \emph{train predictor}). 
We compare these eight existing NAS algorithms with CATE-DNGO: a NAS algorithm  based on CATE encodings with the DNGO subroutine (\emph{train predictor}), and CATE-DNGO-LS: a NAS algorithm based on CATE encodings with the combination of DNGO and LS subroutines (\emph{train predictor}, \emph{perturb arch.}) as described in Section \ref{application}.


\begin{table}[t]

\centering
\begin{adjustbox}{width=1.0\columnwidth}
{\small
\begin{tabular}{@{}l|c|c@{}}
\toprule
\multicolumn{1}{l}{\textbf{NAS methods}} & \multicolumn{1}{c}{\textbf{NAS-Bench-101}} & \multicolumn{1}{c}{\textbf{NAS-Bench-301}}\\
\midrule 
Prev. SOTA \cite{white2019bananas} & 5.92 & 5.35  \\ 
CATE-DNGO-LS (ours) & \textbf{5.88} & \textbf{5.28} \\
\bottomrule
\end{tabular}
}
\end{adjustbox}
\caption{Comparison between CATE and state-of-the-arts: Final test error [\%] given $150$ queried architectures on NAS-Bench-101 and $100$ queried architectures on NAS-Bench-301. The result is averaged over $200$ independent runs.}
\label{table:nasbench101-nasbench301-query-comparison}
\end{table}

\begin{figure}[t]
	\centering
	\includegraphics[width=0.48\textwidth]{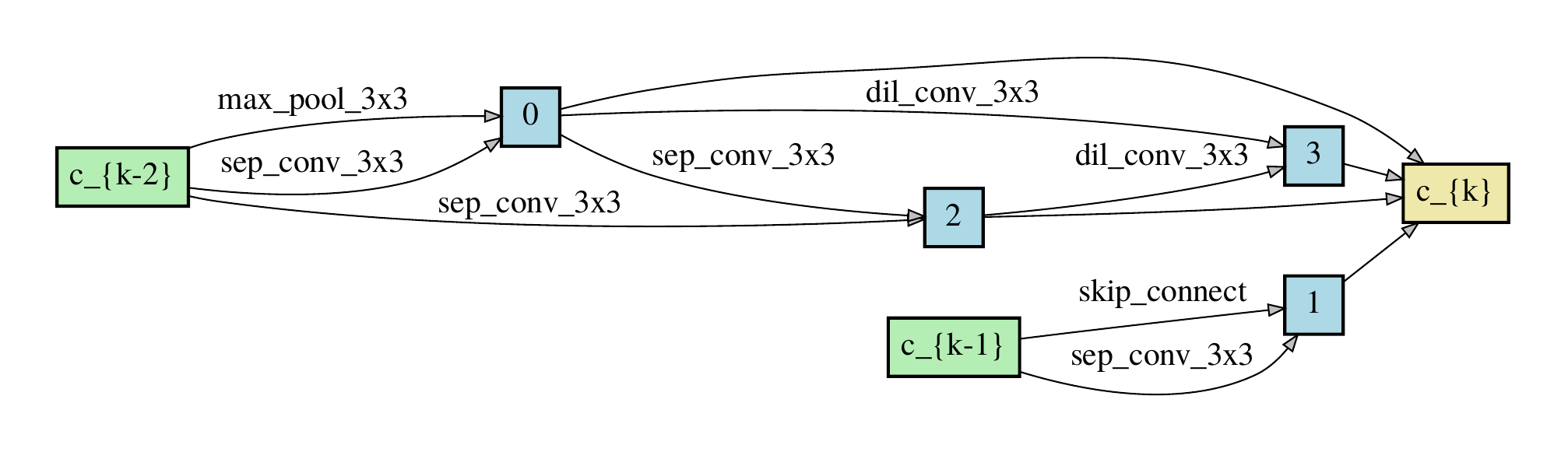}
	\includegraphics[width=0.48\textwidth]{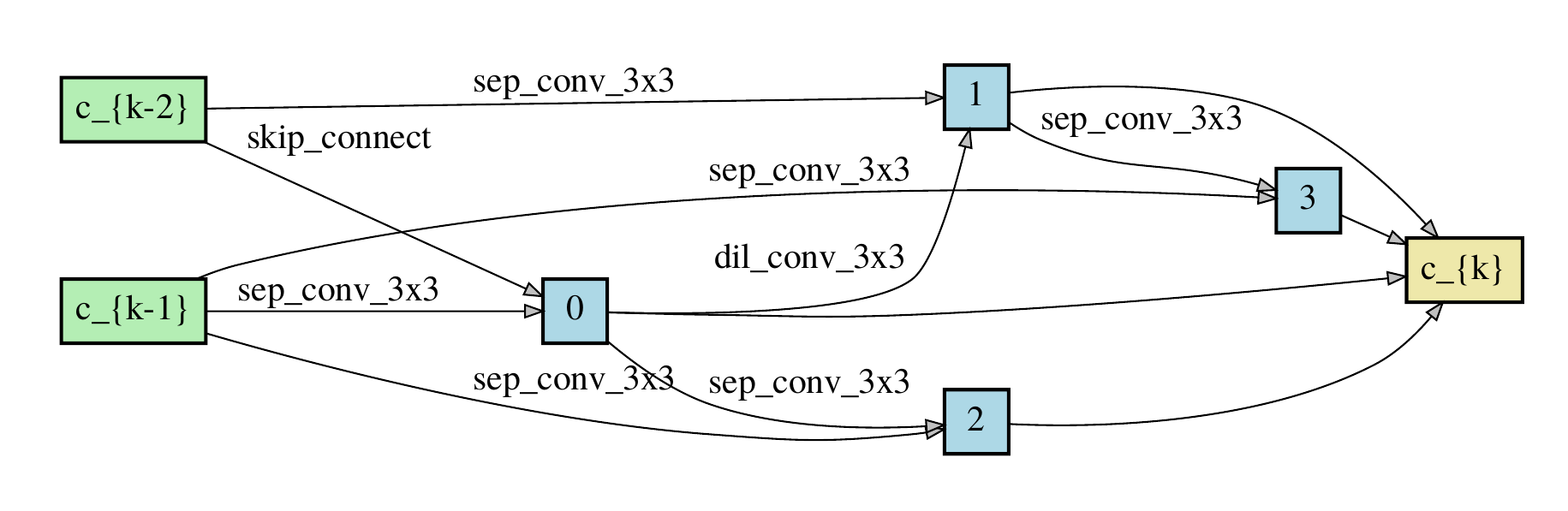}
	\caption{
    Top: Best found cell from CATE-DNGO-LS given the budget of 100 samples. 
	Bottom: Best found cell from CATE-DNGO-LS given the budget of 300 samples. 
	}
	\label{fig:cell}
\end{figure}

Figure \ref{fig.nas_sota_comprison} and Table \ref{table:nasbench101-nasbench301-query-comparison} summarize our results. We have three major findings from Figure \ref{fig.nas_sota_comprison}:
1) Architecture encoding matters especially in the large search space. The right plot shows that CATE-DNGO and CATE-DNGO-LS in DARTS search space not only converge faster but also lead to better final search performance given the same budgets.
2) Local search (LS) is a strong baseline in both small and large search spaces. As mentioned in Section \ref{enc:comp}, performing LS using CATE leads to better results compared to other encodings.
3) NAS algorithms that use more than one encoding-dependent subroutine in general perform better than NAS algorithms with just one subroutine. Specifically, BOGCN and BANANAS that have multiple subroutines perform better than the single-subroutine NAS algorithms such as REA, DNGO, and BOHAMIANN.
Moreover, CATE-DNGO-LS leads to the best performing result in both NAS-Bench-101 and NAS-Bench-301 search spaces.
Meanwhile, the improvement of CATE-DNGO-LS versus CATE-DNGO shrinks in larger search space, indicating that the larger search space is more challenging to encode.
NAS-Bench-301 uses a surrogate model trained on $60$k architectures to predict the performance of all the other architectures in the DARTS search space. The performance of the other architectures, however, can be inaccurate. Given that, we further validate the effectiveness of CATE-DNGO-LS in the \textit{actual} DARTS search space by training the queried architectures from scratch. We set the budget to $100$ and $300$ queries, separately. Each queried architecture is trained for $50$ epochs with a batch size of $96$, using $32$ initial channels and $8$ cell layers. The average validation error of the last $5$ epochs is computed as the label. These values are chosen to be close to the proxy model used in DARTS. It takes about $3.3$ GPU days to finish the search with $100$ quries and $10.3$ GPU days with $300$ queries. See Figure \ref{fig:cell} for the best found cells. To ensure fair comparison, we compare CATE-DNGO-LS to methods \cite{liu2018darts,Li2019RandomSA,yan2020arch,white2019bananas} that use the common test evaluation script which is to train for $600$ epochs with cutout and auxiliary tower. 

Table \ref{table:cifar10-comparison} summarizes our results.
As shown, CATE-DNGO-LS (small budget) achieves competitive performance ($2.55\%$ avg. test error) with much less search cost and CATE-DNGO-LS (large budget) achieves superior performance ($2.46\%$ avg. test error) with similar search cost compared to other sampling-based search methods \cite{yan2020arch,white2019bananas} in the actual DARTS search space. This is consistent with our observation in NAS-Bench-301. We report the transfer learning results on ImageNet \cite{imagenet_cvpr09} in Table \ref{table:imagenet-comparison}. 


\begin{table}[t]
\centering
\begin{adjustbox}{width=1.0\columnwidth}
{\small
\begin{tabular}{@{}l|c|c|c@{}}
\toprule
\multicolumn{1}{l}{\textbf{NAS Methods}} & \multicolumn{1}{c}{\textbf{Avg. Test Error}} & \multicolumn{1}{c}{\textbf{Params}} & \multicolumn{1}{c}{\textbf{Search Cost}}\\
\multicolumn{1}{l}{\textbf{}} & \multicolumn{1}{c}{\textbf{(\%)}} & \multicolumn{1}{c}{\textbf{(M)}} & \multicolumn{1}{c}{\textbf{(GPU days)}}\\
\midrule 
RS \cite{Li2019RandomSA}  & 3.29 $\pm$ 0.15 & \textbf{3.2} & 4  \\ 
DARTS \cite{liu2018darts}  & 2.76 $\pm$ 0.09 & 3.3 & 4  \\ 
BANANAS \cite{white2019bananas}  & 2.67 $\pm$ 0.07 & 3.6 & 11.8 \\ 
arch2vec-BO \cite{yan2020arch} & 2.56 $\pm$ 0.05 & 3.6 & 9.2 \\
\midrule
CATE-DNGO-LS (small budget) & 2.55 $\pm$ 0.08 & 3.5 &  \textbf{3.3} \\
CATE-DNGO-LS (large budget) & \textbf{2.46 $\pm$ 0.05} & 4.1 &  10.3 \\ 
\bottomrule
\end{tabular}
}
\end{adjustbox}
\caption{NAS results in DARTS search space using CIFAR-10.}
\label{table:cifar10-comparison}
\end{table}

\begin{table}[t]
\centering
\begin{adjustbox}{width=1.0\columnwidth}
{\small
\begin{tabular}{@{}l|c|c|c@{}}
\toprule
\multicolumn{1}{l}{\textbf{NAS Methods}} & \multicolumn{1}{c}{\textbf{Params}} & \multicolumn{1}{c}{\textbf{Mult-Adds}} & \multicolumn{1}{c}{\textbf{Top-1 Test Error}}\\
\multicolumn{1}{l}{\textbf{}} & 
\multicolumn{1}{c}{\textbf{(M)}} & \multicolumn{1}{c}{\textbf{(M)}} & \multicolumn{1}{c}{\textbf{(\%)}}\\
\midrule 
SNAS \cite{Xie2018SNAS}  & 4.3 & 522 & 27.3  \\ 
DARTS \cite{liu2018darts} & 4.7 & 574 & 26.7 \\
BayesNAS \cite{pmlr-v97-zhou19e} & \textbf{4.0} & \textbf{440} & 26.5 \\
\textit{arch2vec}-BO \cite{yan2020arch} & 5.2 & 580 & 25.5  \\ 
BANANAS (ours) & 5.1 & 576 & 26.3  \\
\midrule
CATE-DNGO-LS (small budget) & 5.0 & 556 & 26.1 \\  
CATE-DNGO-LS (large budget) & 5.8 & 642 & \textbf{25.0} \\  
\bottomrule
\end{tabular}
}
\end{adjustbox}
\caption{Transfer learning results on ImageNet.}
\label{table:imagenet-comparison}
\end{table}


\subsection{Generalization to Outside Search Space}
In our third experiment, inspired by \cite{white2020study}, we evaluate the generalization ability of CATE beyond the search space on which it was trained. The training search space is designed as a subset of NAS-Bench-101, where each included architecture has 2 to 6 nodes and 1 to 7 edges. The test search space is disjointed from the training search space and includes architectures with 6 nodes and 7 to 9 edges. There are $10,026$ and $60,669$ non-isomorphic graphs in the training and test space respectively. The CATE encodings are pre-trained using the training space and are used to conduct architecture search in the test space. We compare CATE with the adjacency matrix encoding because it was shown in \cite{white2020study} to have \emph{the best generalization capability} compared to other encodings. A simple 2-layer MLP with hidden size $128$ is used as the neural predictor for both encodings. 

\begin{figure}[t]
	\centering
	\includegraphics[width=0.45\textwidth]{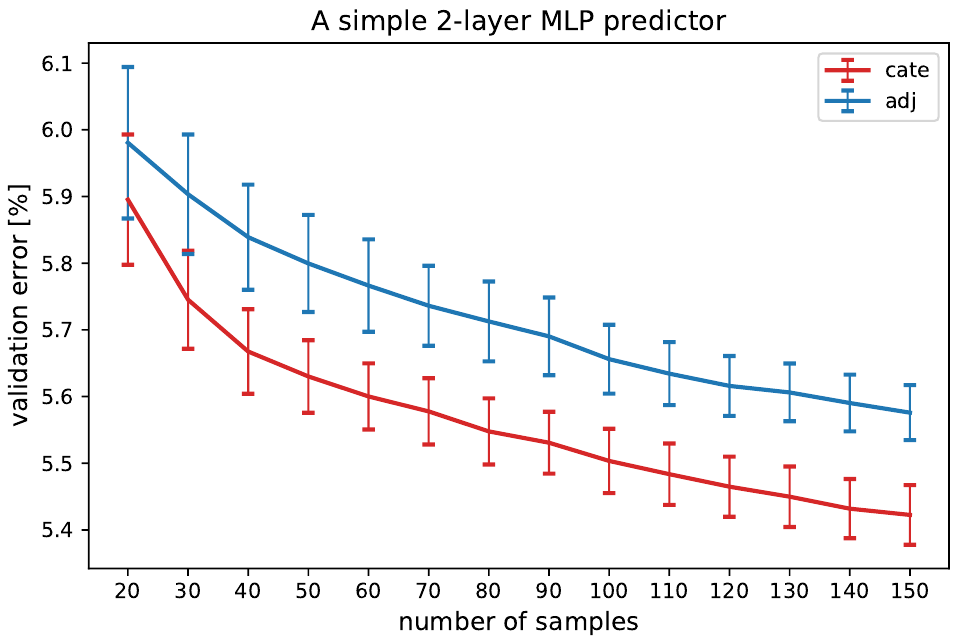}
	\caption{Performance on the out-of-training search space. It reports the validation error of $500$ independent runs.}
	\label{fig.oo}
\end{figure}

Figure \ref{fig.oo} shows the validation error curve of the test search space given the number of $150$ sample budget across $500$ independent runs. As shown, CATE outperforms adjacency matrix encoding by a large margin.
This indicates that CATE can better contextualize the computation information compared to fixed encodings, which generalizes better when adapting to outside search space. Moreover, the padding scheme in our encoder allows us to handle architectures with different numbers of nodes.

\subsection{Ablation Studies}
\label{sec:ablation}
Finally, we conduct ablation studies on different hyperparameters involved in CATE. We use CATE-DNGO as the NAS method and report the final NAS test error [\%] given $150$ queried architectures on NAS-Bench-101. The result is averaged over $200$ independent runs.

\begin{figure}[t]
	\centering
	\includegraphics[width=0.47\textwidth]{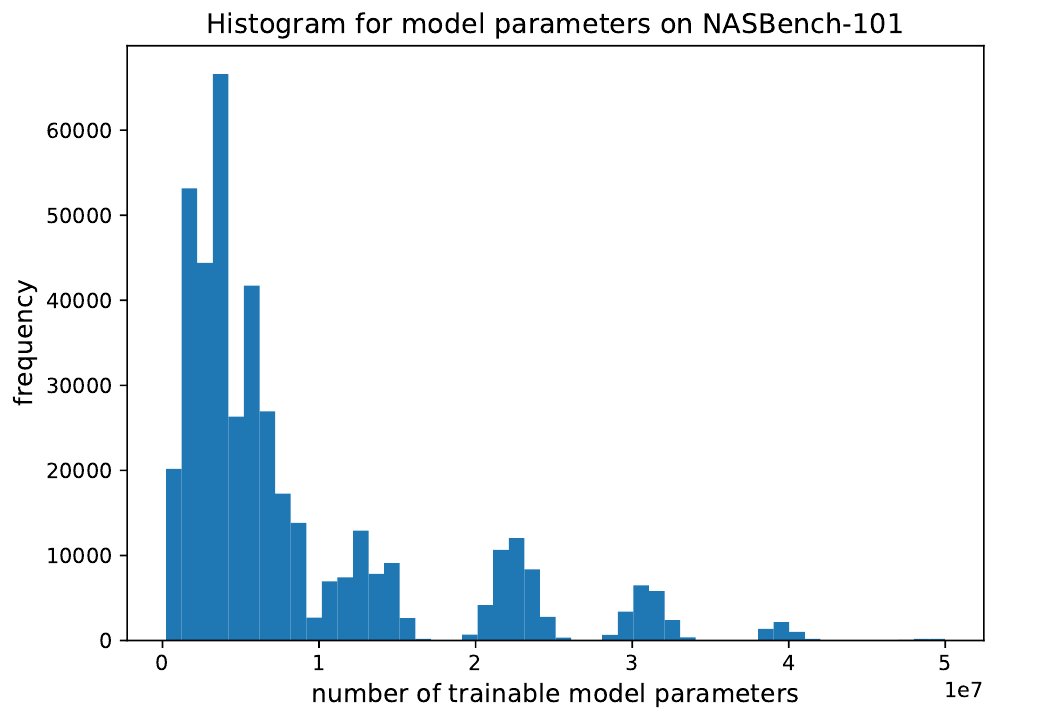}
	\caption{Histogram of model parameters on NAS-Bench-101.}
	\label{fig.histogram}
\end{figure}

\textbf{Architecture Pair Sampling Hyperparameters.}
We plot the histogram of model parameters on NAS-Bench-101 in Figure \ref{fig.histogram}. As shown, the architectures are neither normally nor uniformly distributed in this search space in terms of model parameters. This motivates us to use a sliding window-based architecture pair selection to avoid the unbalanced sampling as proposed in Section \ref{pretraining}. The choice of $\delta$ and $K$ and their effects on the downstream NAS are summarized in Table \ref{table:pair_sampling_choice}. 
We found that strong computation locality (\emph{i.e.} small $\delta$) usually leads to better results. The choice of neighborhood size $K$ does not have a significant effect on NAS performance. Therefore, we choose small $K$ for faster pretraining. For NAS-Bench-301, we use the FLOPs as the computational attributes $\mathbf{P}$ and observe the same trend as in NAS-Bench-101 on the selection of $\delta$ and $K$. We report the results in Appendix \ref{ablation:hp}.

\begin{table}[t] 
\begin{adjustbox}{width=0.76\columnwidth,center}
\scriptsize{
\begin{tabular}{c|c|c|c|c} 
\hline
\diagbox[]{$\delta$}{K} & 1 & 2 & 4 & 8 \\ \hline
$1\times10^{6}$  & 6.02 & 5.95 & 5.99 & 5.95  \\ 
$2\times10^{6}$ & 6.02 & \textbf{5.94} & 6.04 & 5.96  \\
$4\times10^{6}$  & \textbf{5.94} & 6.03 & 6.05 & 5.99  \\
$8\times10^{6}$  & 6.05 & 6.04 & 6.11 & 6.04  \\
\hline
\end{tabular} 
}
\end{adjustbox}
\caption{Effects of $\delta$ and $K$ on architecture pair sampling.}
\label{table:pair_sampling_choice}
\end{table}

\textbf{Transformer Hyperparameters.} We studied the effect of the number of cross-attention Transformer blocks $L_c$ and the hidden dimension of the feed-forward layer $d_{ff}$ on CATE. We fix $\delta$ and $K$ for pre-training as mentioned in Section \ref{sec.experiments}. The downstream NAS result is summarized in Table \ref{table:transformer_choice}. It shows that larger $L_c$ and $d_{ff}$ usually lead to better NAS performance, which indicates that deep contextualized representations are beneficial to downstream NAS.  

\begin{table}[ht] 
\begin{adjustbox}{width=0.71\columnwidth,center}
\scriptsize{
\begin{tabular}{c|c|c|c} 
\hline
\diagbox[]{$d_{ff}$}{$L_c$} & 6 & 12 & 24 \\ \hline
64  & 6.07 &  5.99 & 5.95  \\ 
128 & 6.01 & \textbf{5.94} & 5.95   \\
256  & 5.97 & \textbf{5.94} & \textbf{5.94}  \\
\hline
\end{tabular} 
}
\end{adjustbox}
\caption{Effects of $L_c$ and $d_{ff}$ on pretraining CATE.}
\label{table:transformer_choice}
\end{table}

\textbf{Choice of Mask Type.} We studied pretraining CATE with direct/indirect dependency mask and summarize its downstream NAS results in Table \ref{table:mask_type}. CATE trained with indirect dependency mask outperforms the direct one in both benchmarks, indicating that capturing long-range dependency helps preserve computation information in the encodings.


\begin{table}[ht]
\centering
\begin{adjustbox}{width=0.81\columnwidth}
{\small
\begin{tabular}{@{}c|c|c@{}}
\toprule
\multicolumn{1}{c}{\textbf{Mask type}} & \multicolumn{1}{c}{\textbf{NAS-Bench-101}} & \multicolumn{1}{c}{\textbf{NAS-Bench-301}} \\ 
\midrule 
Direct  & 6.03 & 5.35  \\ 
Indirect  & \textbf{5.94} & \textbf{5.30}  \\ 
\bottomrule
\end{tabular}
}
\end{adjustbox}
\caption{Direct/Indirect dependency mask selection.}
\label{table:mask_type}
\end{table} 
\section{Conclusion}
\label{sec.conclusion}
 In this paper, we presented CATE, a new computation-aware architecture encoding method based on Transformers. Unlike encodings with fixed transformations, we show that the computation information of neural architectures can be contextualized through a pairwise learning scheme trained with MLM. Our experimental results show its effectiveness and scalability along with three major encoding-dependent NAS subroutines in both small and large search spaces. We also show its superior generalization capability outside the training search space. We anticipate that the methods presented in this work can be extended to encode even larger search spaces (\emph{e.g.} TuNAS \cite{bender2020weight}) to study the effectiveness of different downstream NAS algorithms. 

\section*{Acknowledgement}
\label{sec.ack}
%
%
We would like to thank the anonymous reviewers for their helpful comments. We thank Yu Zheng, Colin White, and Frank Hutter for their help with this project. This work was partially supported by NSF Awards CNS-1617627, CNS-1814551, and PFI:BIC-1632051.
\bibliography{icml2021}
\bibliographystyle{icml2021}

\clearpage
\appendix

\section{Uni/Bidirectional Encoding} \label{uni-bi-enc}
As mentioned in Section \ref{enc}, we also considered encoding the architecture in a bidirectional manner where both the output node hidden vector from the original DAG and the input node hidden vector from the reversed one are extracted and then concatenated together. Note that $d_{c}$ in the cross-attention Transformer encoder will be doubled due to the concatenation. We compare the results of unidirectional and bidirectional encodings  in Table \ref{table:encoding_type}. As shown, bidirectional encoding does not necessarily improve the results. Therefore, we keep unidirectional encoding in other experiments due to its simplicity and better performance.


\begin{table}[ht]
\centering
\begin{adjustbox}{width=0.95\columnwidth}
{\small
\begin{tabular}{@{}l|c|c@{}}
\toprule
\multicolumn{1}{l}{\textbf{Encoding}} & \multicolumn{1}{c}{\textbf{NAS-Bench-101}} & \multicolumn{1}{c}{\textbf{NAS-Bench-301}} \\ 
\midrule 
Unidirectional & \textbf{5.88} & \textbf{5.28} \\ 
Bidirectional & 5.89 &  5.30 \\ 
\bottomrule
\end{tabular}
}
\end{adjustbox}
\caption{Unidirectional encoding vs. bidirectional encoding. We report the final NAS test error [\%] given 150 queried architectures on NAS-Bench-101 and 100 queried architectures on NAS-Bench-301. The result is averaged over 200 independent runs.}
\label{table:encoding_type}
\end{table}

\section{Architecture Pair Sampling Hyperparameters} \label{ablation:hp}

\begin{figure}[ht]
	\centering
	\includegraphics[width=0.47\textwidth]{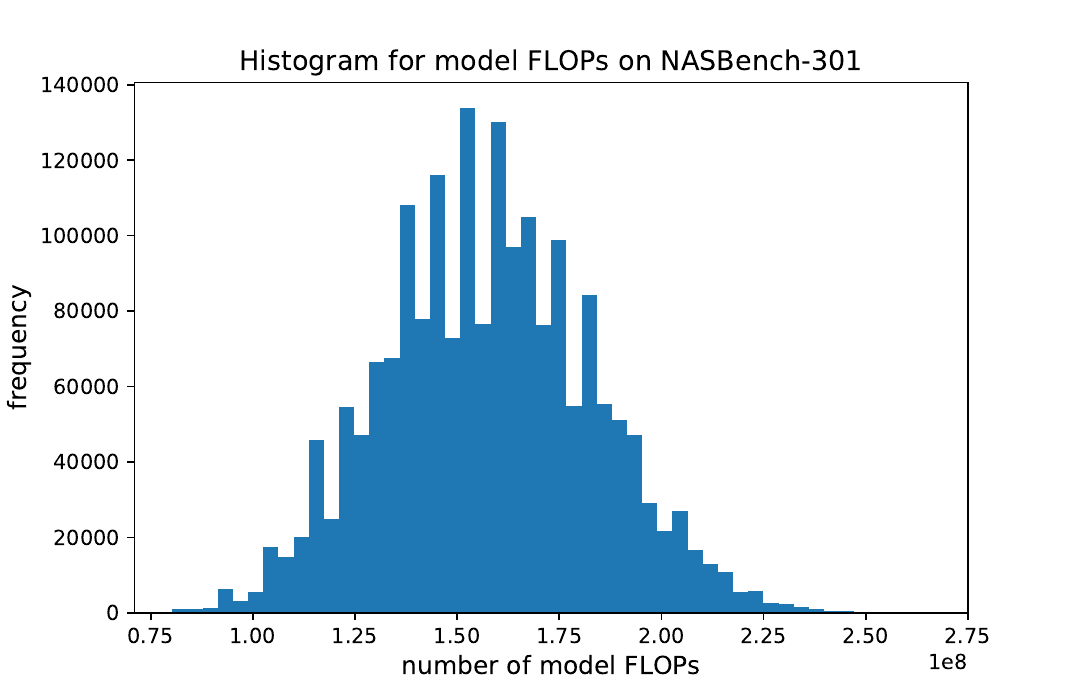}
	\caption{Histogram of model FLOPs on the sampled $1,000,000$ architectures of NAS-Bench-301.}
	\label{fig.hist-nas301-flops}
\end{figure}

As mentioned in Section \ref{sec:ablation}, we randomly sample 1,000,000 architectures in NAS-Bench-301 for pretraining. We use the same proxy model configuration (\emph{i.e.} 100 training epochs, 32 initial channels, 8 cell layers) as used in NAS-Bench-301 to compute the model FLOPs. We plot the histogram of model FLOPs of the sampled architectures in Figure \ref{fig.hist-nas301-flops}. Given that, we experiment different $\delta$ and $K$ and summarize the downstream NAS results in Table \ref{table:pair_sampling_choice_301}. Similar to our reported results on NAS-Bench-101, we find that strong locality leads to better results.

\begin{table}[ht] 
\begin{adjustbox}{width=0.82\columnwidth,center}
\scriptsize{
\begin{tabular}{c|c|c|c|c} 
\hline
\diagbox[]{$\delta$}{K} & 1 & 2 & 4 & 8 \\ \hline
$5\times10^{6}$  & \textbf{5.28} & 5.29 & 5.30 & 5.30  \\ 
$1\times10^{7}$ & 5.30 & \textbf{5.28} & 5.29 & 5.31  \\
$2\times10^{7}$ & 5.30 & 5.30 & 5.31 & 5.32  \\
\hline
\end{tabular} 
}
\end{adjustbox}
\caption{Effects of $\delta$ and $K$ on architecture pair sampling on NAS-Bench-301. We report the final NAS test error [\%] given 100 queried architectures on NAS-Bench-301. The result is averaged over 200 independent runs.}
\label{table:pair_sampling_choice_301}
\end{table}

\section{Corruption Rate}
By default, we randomly select 20\% operations from each architecture within the pair for masking in the pairwise pre-training. We also experimented corruption rates of 15\% and 30\%. As shown in Table \ref{table:corruption_rate}, overall, we find that the corruption rate has a limited effect on the NAS performance. Note that the number of nodes in our search space is much smaller compared to the number of tokens in the sequence modeling tasks. Given that, using larger corruption rate may slow down the training convergence and result in degraded performance. Based on these results, we use 20\% corruption rate for other experiments.

\begin{table}[t]
\centering
\begin{adjustbox}{width=0.96\columnwidth}
{\small
\begin{tabular}{@{}c|c|c@{}}
\toprule
\multicolumn{1}{c}{\textbf{Corruption Rate}} & \multicolumn{1}{c}{\textbf{NAS-Bench-101}} & \multicolumn{1}{c}{\textbf{NAS-Bench-301}} \\ 
\midrule 
15\% & 5.89 & \textbf{5.28}   \\
20\% & \textbf{5.88} &  \textbf{5.28} \\
30\% & 5.93 & 5.29 \\
\bottomrule
\end{tabular}
}
\end{adjustbox}
\caption{NAS results under different corruption rates.}
\label{table:corruption_rate}
\end{table}

\section{NAS-Bench-301 Benchmark}
\label{sec.301-cell}

\begin{figure*}[ht]
	\centering
	\includegraphics[width=0.86\textwidth]{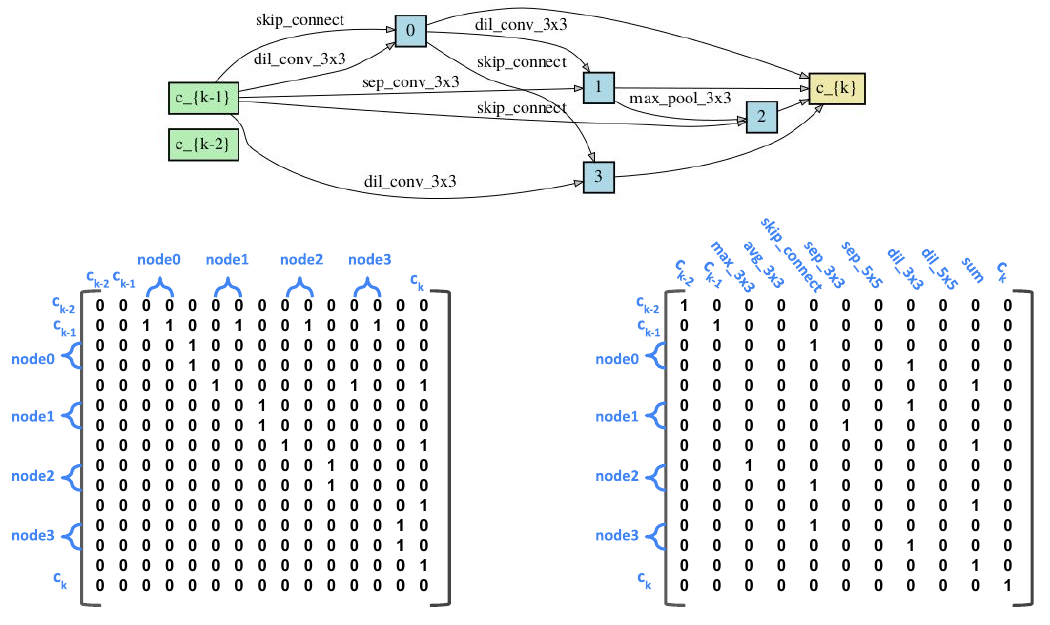}
	\caption{A cell transformation example in DARTS search space. The top  panel  shows  the cell. The bottom-left and bottom-right panels show its corresponding adjacency matrix and operation matrix respectively.}
	\label{fig.cell}
\end{figure*}

NAS-Bench-301 \cite{julien2020nas301} is the first surrogate NAS benchmark to cover the large-scale DARTS search space \cite{liu2018darts}. The DARTS search space consists of two cells: a convolutional cell and a reduction cell, each with six nodes. For each cell, the first two nodes are the outputs from the previous two cells. The next four nodes contain two edges as input, creating a DAG. In total, there are roughly $10^{18}$ DAGs without considering graph isomorphism, which is a much larger search space compared to NAS-Bench-101 \cite{pmlr-v97-ying19a} and NAS-Bench-201 \cite{dong2020nasbench201}.  

NAS-Bench-301 is fully trained on around $60$k architectures collected by unbiased architecture sampling using random search as well as biased and dense architecture sampling in high-performance regions using different NAS methods and training hyperparameters (including training time, number of parameters, and number of multiply-adds). It trains various regression models such as Random Forest (RF) \cite{breiman2001random}, Support Vector Regression (SVR) \cite{NIPS1996_d3890178}, Graph Isomorphism Network (GIN) \cite{xu2018how} and Tree-based gradient boosting model (\emph{e.g.} XGBoost (XGB) \cite{Chen:2016:XST:2939672.2939785}, LGBoost (LGB) \cite{NIPS2017_6449f44a}) to predict the accuracies of unseen architectures. The three best-performing models (GIN, XGB, LGB) are used to predict the search trajectories in the  benchmark API. 

\subsection{Cell Transformation}
To transform the DARTS search space into one with the same input format as NAS-Bench-101, we additionally add a summation node to make nodes to represent operations and edges to represent data flow. For example, if there is an edge from node A to node B with operation O, we create an additional node P, remove the edge $\langle A,B \rangle$, and add 2 edges  $\langle A, P \rangle$ and $\langle P, B \rangle$. The operation on node P is set to be O. Given that, a $15 \times 15$ upper-triangular binary matrix is used to encode edges and a $15 \times 11$ operation matrix is used to encode operations with the order of \{$c_{k-2}$, $c_{k-1}$, 3 $\times$ 3 max-pool, 3 $\times$ 3 average-pool, skip connect, 3 $\times$ 3 separable conv, 5 $\times$ 5 separable conv, 3 $\times$ 3 dilated conv, 5 $\times$ 5 dilated conv, sum, $c_{k}$\}. Following NAS-Bench-301 \cite{julien2020nas301}, we do not include zero operator.   Following \cite{liu2018progressive}, we use the same cell for both normal and reduction cells.
An example of cell transformation is shown in Figure \ref{fig.cell}.



\end{document}